# AN ANALYSIS OF GENE EXPRESSION DATA USING PENALIZED FUZZY C-MEANS APPROACH

P.K. Nizar Banu[1] H. Hannah Inbarani[2]

[1]Department of Computer Applications, B. S. Abdur Rahman University, Chennai, Tamilnadu, India
Email: nizarbanu@gmail.com

[2]Department of Computer Science, Periyar University, Salem, Tamilnadu, India
Email: hhinba@gmail.com

*ABSTRACT*

With the rapid advances of microarray technologies, large amounts of high-dimensional gene expression data are being generated, which poses significant computational challenges. A first step towards addressing this challenge is the use of clustering techniques, which is essential in the data mining process to reveal natural structures and identify interesting patterns in the underlying data. A robust gene expression clustering approach to minimize undesirable clustering is proposed. In this paper, Penalized Fuzzy C-Means (PFCM) Clustering algorithm is described and compared with the most representative off-line clustering techniques: K-Means Clustering, Rough K-Means Clustering and Fuzzy C-Means clustering. These techniques are implemented and tested for a Brain Tumor gene expression Dataset. Analysis of the performance of the proposed approach is presented through qualitative validation experiments. From experimental results, it can be observed that Penalized Fuzzy C-Means algorithm shows a much higher usability than the other projected clustering algorithms used in our comparison study. Significant and promising clustering results are presented using Brain Tumor Gene expression dataset. Thus patterns seen in genome-wide expression experiments can be interpreted as indications of the status of cellular processes. In these clustering results, we find that Penalized Fuzzy C-Means algorithm provides useful information as an aid to diagnosis in oncology.

*Keywords*: Clustering, Microarray, Gene Expression, Brain Tumor, Fuzzy Clustering, FCM, PFCM

## 1.INTRODUCTION

Gene expression is the fundamental link between genotype and phenotype in a species, with microarray technologies facilitating the measurement of thousands of Gene expression values under tightly controlled conditions, e.g. (i) from a particular point in the cell cycle, (ii) after an interval response to some environmental change, (iii) from RNA, isolated from a tissue exhibiting certain phenotypic characteristics and so on (Kerr, et. al., 2008). A problem inherent in the use of microarray technologies is the huge amount of data produced. Searching for meaningful information patterns and dependencies among genes, in order to provide a basis for hypothesis testing, typically includes the initial step of a natural basis for organizing gene expression data to group genes together with similar patterns of expression. The field of gene expression data analysis has grown in the past few years from being purely data-centric to integrative, aiming at complementing microarray analysis with data and knowledge from diverse available sources. Advances in microarray technologies have made it possible to measure the expression profiles of thousands of genes in parallel under varying experimental conditions. Due to the large number of genes and complex gene regulation networks, clustering is a useful exploratory technique for analyzing these data. It divides data of interest into a small number of relatively homogeneous groups or clusters. Clustering is a popular data mining technique for various applications. One of the reasons for its popularity is the ability to work on datasets with minimum or on a priori knowledge. This makes clustering practical for real world applications. We can view the expression levels of different genes as attributes of the samples, or the samples as the attributes of different genes. Clustering can be performed on genes or samples (Michael et. al. 1998). This paper introduces the application of Penalized Fuzzy C-Means algorithm to cluster Brain Tumor genes.

This paper is organized as follows. In Section 2, Research background for clustering Gene Expression patterns is discussed. In Section 3, Methodology for preparing Gene Expression patterns is presented. Section 4 presents a method for extracting highly suppressed and expressed genes based on Penalized Fuzzy C-Means algorithm. The Experimental results are discussed in Section 5. Finally, section 6 concludes this paper by enumerating the merits of the proposed approaches.

## 2. RESEARCH BACKGROUND

### *2.1. Clustering*

Clustering genes, groups similar genes into the same cluster based on a proximity measure. Genes in the same cluster have similar expression patterns. One of the characteristics of gene expression data is that it is meaningful to cluster both genes and samples. The most commonly applied full space clustering algorithms on gene expression profiles are hierarchical clustering algorithms (Michael et. al. 1998), self-organizing maps (Paul, 1999) and K-Means clustering algorithms (Tavazoie, et. al., 1999). Hierarchical algorithms merge genes with the most similar expression profiles iteratively in a bottom-up manner. Self-organizing maps and K-means algorithms partition genes into user-specified *k* optimal clusters. Other full space clustering algorithms applied on gene expression data include Bayesian network (Friedman et. al. 2000) and neural network. A robust image segmentation method that combines the watershed segmentation and penalized fuzzy Hopfield neural network algorithms to minimize over-segmentation is described in (Kuo et. al. 2006). (Brehelin et. al., 2008) evaluates the stability of clusters derived from hierarchical clustering by taking repeated measurements. A wide variety of clustering algorithms are available for clustering gene expression data (Bezdek, 1981). They are mainly classified as Partitioning methods, hierarchical methods, Density based methods, Model based methods, Graph Theoretic methods, soft computing methods etc.

Multiple expression measurements are commonly recorded as a real-valued matrix, with row objects corresponding to gene expression measurements over a number of experiments and columns corresponding to the pattern of expression of all genes for a given microarray experiment. Each entry $x_{ij}$, is the measured expression of gene i in experiment j. Dimensionality of a gene refers to the number of expression values recorded for it. A gene/gene-profile x is a single data item (row) consisting of d measurements, $x = (x_1, x_2, . . . , x_d)$. An experiment/sample y is a single microarray experiment corresponding to a single column in the gene expression matrix, $y = (x_1, x_2, …, x_n)^T$ where n is the number of genes in the dataset. Clustering is considered an interesting approach for finding similarities in data and putting similar data into groups. Initial step in the analysis of gene expression data is the detection of groups of genes that exhibit similar expression patterns. In gene expression, elements are usually genes and the vector of each gene is its expression pattern. Patterns that are similar are allocated in the same cluster, while the patterns that differ significantly are put in different clusters. Gene expression data are usually of high dimensions and relatively small samples, which results in the main difficulty for the application of clustering algorithms. Clustering the microarray matrix can be achieved in two ways:

(i) Genes can form a group which show similar expression across conditions,
(ii) Samples can form a group which shows similar gene expression across all genes.

This gives rise to global clustering, where a gene or sample is grouped across all dimensions. Additionally, the clustering can be complete or partial. A complete clustering assigns each gene to a cluster, whereas a partial clustering does not. Partial clustering tends to be more suited to gene expression, as the dataset often contains irrelevant genes or samples. Clearly this allows:

(i) Noisy genes to be left out, with correspondingly less impact on the outcome and
(ii) Genes belonging to no cluster - omitting a large number of irrelevant contributions.

Microarrays measure expression for the entire genome in one experiment, but genes may change expression, independent of the experimental condition. Forced inclusion in well-defined but inappropriate groups may impact the final structures found for the data. Partial clustering avoids the situation where an interesting sub-group in a cluster is hidden through forcing membership of unrelated genes (Kerr, et. al., 2008).

## *2.2. Categories of Gene Expression data clustering*

Methods of clustering can be categorized as Hard Clustering or Soft Clustering. Hard Clustering requires each gene to belong to a single cluster, whereas Soft Clustering permit genes to simultaneously be members of numerous clusters. Hard Clustering tells whether a gene belongs to a cluster or not. Whereas in Soft Clustering, with membership values, every gene belongs to each cluster with a membership weight between 0 (doesn't belong) and 1 (belongs). Clustering algorithms which permit genes to belong to more than one cluster are more applicable to Gene expression. Gene expression data has certain special characteristics and is a challenging research problem

A modern working definition of a gene is "a locatable region of genomic sequence, corresponding to a unit of inheritance, which is associated with regulatory regions, transcribed regions, and or other functional sequence regions". Currently, a typical microarray experiment contains $10^3$ to $10^4$ genes, and this number is expected to reach to the order of $10^6$. However, the number of samples involved in a microarray experiment is generally less than 100. One of the characteristics of gene expression data is that it is meaningful to cluster both genes and samples. On one hand, co-expressed genes can be grouped into clusters based on their expression patterns (Ben-Dor, et. al., 1999 & Michael et. al. 1998). In such gene-based clustering, the genes are treated as the objects, while the samples are the features. On the other hand, the samples can be partitioned into homogeneous groups. Each group may correspond to some particular macroscopic phenotype, such as clinical syndromes or cancer types (Golub et. al., 1999). Such sample-based clustering considers the samples as the objects and the genes as the features. The distinction of gene-based clustering and sample-based clustering is based on different characteristics of clustering tasks for gene expression data. Some clustering algorithms, such as K-Means and hierarchical approaches, can be used both to group genes and to partition samples.

Hard clustering algorithms like K-Means and k-medoid place a restriction that a data object can belong precisely to only one cluster during clustering process. This can be too restrictive while clustering high dimensional data like Gene Expression data because genes have a property of getting expressed in multiple conditions. Fuzzy set clustering like Fuzzy C-Means, Penalized Fuzzy C-Means allows data objects to belong to multiple clusters based on the degree of membership.

### 2.3. Analysis of Gene Expression data

*Gene expression data* is usually represented by a matrix, with rows corresponding to genes, and columns corresponding to conditions, experiments or time points. The content of the matrix is the expression levels of each gene under each condition. Those levels may be absolute, relative or otherwise normalized. Each column contains the results obtained from a single array in a particular condition, and is called the *profile* of that condition. Each row vector is the *expression pattern* of a particular gene across all the conditions.

Analyzing gene expression data is a process by which a gene's information is converted into the structures and functions of a cell. Thousands of different mRNAs are present in a given cell; together they make up the transcriptional profile. It is important to remember that when a gene expression profile is analyzed in a given sample, it is just a snapshot in time and space.

### 2.4 Clustering Techniques

#### K-Means Algorithm

The K-means method aims to minimize the sum of squared distances between all points and the cluster centre. This procedure includes the steps, as described by Tou and Gonzalez (Tou et. al. 1974).

#### Rough Set Theory

Rough set theory introduced by Pawlak (Pawlak, 1982) deals with uncertainty and vagueness. Rough set theory became popular among scientists around the world due to its fundamental importance in the field of artificial intelligence and cognitive sciences. Similar to fuzzy set theory it is not an alternative to classical set theory but it is embedded in it. Rough set theory can be viewed as a specific implementation of Frege's idea of vagueness, i.e., imprecision in this approach is expressed by a boundary region of a set, and not by a partial membership, like in fuzzy set.

#### Rough Clustering

A rough cluster is defined in a similar manner to a rough set that is with lower and upper approximation. The lower approximation of a rough cluster contains genes that only belong to that cluster. The upper approximation of a rough cluster contains genes in the cluster which are also members of other clusters (SushmitaMitra, 2004 & SushmitaMitra, 2006). To use the theory of rough sets in clustering, the value set ($V_a$) need to be ordered. This allows a measure of the distance between each object to be defined. Distance is a form of similarity, which is a relaxing of the strict requirement of indiscernibility outlined in canonical rough sets theory, and allows the inclusion of genes that are similar rather than identical. Clusters of genes are then formed on the basis of their distance from each other. An important distinction between rough clustering and traditional clustering approaches is that, with rough clustering, an object can belong to more than one cluster.

*Fuzzy Clustering*

Cluster analysis is a method of grouping data with similar characteristics into larger units of analysis. First in (Zadeh, 1965) fuzzy set theory that gave rise to the concept of partial membership, based on a membership function, fuzziness was articulated and has received increasing attention. Fuzzy clustering which produces overlapping cluster partitions has been widely studied and applied in various areas. In fuzzy clustering, the Fuzzy C-Means (FCM) clustering algorithm is the best known and most powerful methods used in cluster analysis (Bezdek, 1981). In (Yu et. al., 2007), a general theoretical method to evaluate the performance of fuzzy clustering algorithm is proposed. The Fuzzy integrated model is accurate than rough integrated model and conventional integrated model (Banu et. al., 2011). Fuzzy clustering approach captures the uncertainty that prevails in gene expression and becomes more suitable for tumor prediction. One of the important parameters in the FCM is the weighting exponent *m*. When *m* is close to one, the FCM approaches the hard C-Means algorithm. When *m* approaches infinity, the only solution of the FCM will be the mass center of the data set. Therefore, the weighting exponent *m* plays an important role in the FCM algorithm.

## 3. METHODOLOGY

Cluster analysis, is an important tool in gene expression data analysis. For experimentation, we used a set of gene expression data that contains a series of gene expression measurements of the transcript (mRNA) levels of brain tumor gene. In clustering gene expression data, the genes are treated as objects and the samples are treated as attributes. Gene pattern extraction consists of 4 steps.
i) Data Preparation
ii) Data Normalization
iii) Clustering
iv) Pattern analysis

### *3.1. Data Preparation*

We represent the gene expression data as $n_g$ by $n_s$ matrix: $\hat{M} = \{m_{i,j} | i=1,2,..n_g, j=1,2,..n_s\}$. There are $n_s$ columns, one for each sample and $n_g$ rows, one for each gene. One row of genes is also called a gene vector, denoted as $\vec{g}_i = \langle m_{i,1}, m_{i,2}, ..., m_{i,n_s} \rangle$. Thus a gene vector contains the values of a particular attribute for all samples.

### *3.2. Data Normalization*

Data sometimes need to be transformed before being used. For example; attributes may be measured using different scales, such as centimeters and kilograms. In instances where the range of values differ widely from attribute to attribute, these differing attribute scales can dominate the results of the cluster analysis. It is therefore common to normalize the data so that all attributes are on the same scale. The following are two common approaches for data normalization of each gene vector:

$$m'_{i,j} = \frac{m_{i,j} - \overline{m}_i}{\overline{m}_i},$$

or

$$m'_{i,j} = \frac{m_{i,j} - \overline{m}_i}{\sigma_i},$$

where

$$\overline{m}_i = \frac{\sum_{j=1}^{n_s} m_{i,j}}{n_s}, \sigma_i = \frac{\sqrt{\sum_{j=1}^{n_s}(m_{i,j}-\overline{m}_i)^2}}{n_s - 1}$$

and $m'_{i,j}$ denotes the normalized value for gene vector *i* of sample *j*, $m_{i,j}$ represents the original value for gene *i* of sample *j*, $n_s$ is the number of samples, $\overline{m}_i$ is the mean of the values for gene vector *i* over all samples, and $\sigma_i$ is the standard deviation of the $i^{th}$ gene vector.

The Brain Tumor gene expression data is used for our experiments. This data set is publically available in Broad Institute web site. The various cluster validation techniques namely Root Mean Square Error (RMSE), Mean Absolute Error (MAE) and Xie-Beni (XB) validity index are used to validate the clusters obtained after applying the clustering algorithms.

## 4. PROPOSED PPROACH: PENALIZED FUZZY C-MEANS

Penalized Fuzzy C-Means (PFCM) algorithm for clustering gene expression data is introduced in this paper, which modified Fuzzy C-Means (FCM) algorithm to produce more meaningful fuzzy clusters. Genes are assigned a membership degree to a cluster indicating its percentage association with that cluster. The two algorithms differ in the weighting scheme used for the contribution of a gene to the mean of the cluster. FCM membership values for a gene are divided among clusters in proportion to similarity with that clusters mean. The contribution of each gene to the mean of a cluster is weighted, based on its membership grade. Membership values are adjusted iteratively until the variance of the system falls below a threshold. These calculations require the specification of a degree of fuzziness parameter which is problem specific (Dembele et. al., 2003).

The membership weighting system reduces noise effects, as a low membership grade is less important in centroid calculation.

PFCM algorithm helps in identifying hidden pattern and providing enhanced understanding of the functional genomics in a better way.

Fuzzy Clustering permit genes to belong to more than one cluster, is more applicable to Gene Expression. Noisy genes are unlikely to be members of several clusters and genes with similar change in expression for a set of samples are involved in several biological functions and groups should not be co-active under all conditions. This gives rise to high inconsistency in the gene groups and some overlap between them. The boundary of a cluster is usually fuzzy for three reasons:

i. The gene expression dataset might be noisy and incomplete
ii. The similarity measurement between genes is continuous and there is no clear cutoff value for group membership
iii. A gene might behave similarly to gene1 under a set of samples and behave similarly to another gene2 under another set of samples. Therefore, there is a great need for a fuzzy clustering method, which produces clusters in which genes can belong to a cluster partially and to multiple clusters at the same time with different membership degrees.

The main objective of using this method is to minimize the objective function value so that the highly suppressed genes and highly expressed genes are clustered separately and also it helps to diagnose at an early stage of tumor formation.

### 4.1. Penalized Fuzzy C-Means Algorithm

Another strategy for fuzzy clustering, called the penalized Fuzzy C-Means (PFCM) algorithm, with the addition of a penalty term was proposed by Yang (Yang, 1993 & Yang, 1994). Yang made the fuzzy extension of the Classification Maximum Likelihood (CML) procedure in conjunction with fuzzy C-partitions and called it a class of fuzzy CML procedures. The idea of penalization is also important in statistical learning. Combining the CML procedure and penalty idea, Yang (Yang, 1993) added a penalty term to the FCM objective function $J_{FCM}$ and then extended the FCM to the so-called Penalized FCM (PFCM) which produces more meaningful and effective results than the FCM algorithm. Thus, the PFCM objective function is defined as follows:

$$J_{PFCM} = \frac{1}{2}\sum_{j=1}^{c}\sum_{i=1}^{n} u_{i,j}^m \|x_i - w_j\|^2 - \frac{1}{2}v\sum_{j=1}^{c}\sum_{i=1}^{n} u_{i,j}^m \ln\alpha_j$$

$$= J_{FCM} - \frac{1}{2}v\sum_{j=1}^{c}\sum_{i=1}^{n} u_{i,j}^m \ln\alpha_j$$

where $\alpha_j$ is a proportional constant of class j and v ($\geq 0$) is a constant. The penalty term $-\frac{1}{2}v\sum_{j=1}^{c}\sum_{i=1}^{n} u_{i,j}^m \ln\alpha_j$ is added to the objective function, when v=0, $J_{PFCM}$ is equal to $J_{FCM}$.

$\alpha_j$, $w_j$ and $u_{i,j}$ are defined as

$$\alpha_j = \frac{\sum_{i=1}^{n} u_{i,j}^m}{\sum_{j=1}^{c}\sum_{i=1}^{n} u_{i,j}^m}, \; j=1,2,...,c \quad (1)$$

$$w_j = \frac{1}{\sum_{i=1}^{n}(u_{i,j})^m}\sum_{i=1}^{n}(u_{i,j})^m x_i \quad (2)$$

$$u_{i,j} = \left(\sum_{l=1}^{c}\frac{(\|x_i - w_j\|^2 - v\ln\alpha_j)^{1/(m-1)}}{(\|x_i - w_l\|^2 - v\ln\alpha_l)^{1/(m-1)}}\right)^{-1} \quad (3)$$

Based on the numerical results PFCM is more accurate than FCM.
The steps of the PFCM algorithm are given as follows:

Step 1: Initialize the cluster centroids $w_j(2\leq j \leq c), v(v>0)$, fuzzification parameter, $m(1\leq m<\infty)$, and the value $\varepsilon > 0$. Gives a fuzzy C-partition $\Omega^{(0)}$ and t=1.

Step 2: Calculate $\alpha_j^{(t)}, w_j^{(t)}$ with $\Omega^{(t-1)}$ using Eqs. (1) and (2).

Step 3: Calculate the membership matrix
$\Omega^{(t)} = [u_{i,j}]$ with $\alpha_j^{(t)}, w_j^{(t)}$
using Eq. (3)

Step 4: Compute $\Delta = \max(|\Omega^{(t+1)} - \Omega^{(t)}|)$.
If $\Delta > \varepsilon$, and, $t = t+1$ go to Step 2; otherwise go to Step 5.

Step 5: Find the results for the final class centroids.

## 5. EXPERIMENTAL RESULTS

The effectiveness of the algorithms based on cluster validity measure is demonstrated in this section.

### 5.1. Data Source

A set of gene expression data that contains a series of gene expression measurements of the transcript (mRNA) levels of brain Tumor gene is used in this paper to analyze the efficiency of the proposed approach.

The brain tumor dataset is taken from the Broad-Institute website (http://www.broadinstitute.org/cgi-bin/cancer/datasets.cgi). The dataset is titled as *"Gene Expression-Based Classification and Outcome Prediction of Central Nervous*

*System Embryonal Tumors"*. The Datasets Consists of three types of Brain Tumors data namely Medulloblastoma classic and desmoplastic, Multiple Brain tumors, Medulloblastoma treatment outcome. Each dataset contains nearly 7000 genes with 42 samples.

In order to evaluate the proposed algorithm, we applied it to the Brain Tumor gene expression data taken from Broad Institute by taking 7129, 5000, 3000, and 1000 genes for all samples with various numbers of clusters.

### 5.2. Comparative Analysis

A comparative study of the performance of K-Means (Tou et. al. 1974), Rough K-Means (Peters, 2006) and Fuzzy C-Means (Peters, 2006) is made with Penalized Fuzzy C-Means algorithm (Shen et. al, 2006).

*Cluster Validation*

In this paper, Root Mean Square Error, Mean Absolute Error and Xie-Beni validity index are used to validate the clusters obtained after applying the clustering algorithms. To assess the quality of our method, we need an objective external criterion. In order to validate our clustering results, we employed Root Mean Square Error (RMSE), Mean Absolute Error (MAE) (Pablo de Castro et. al., 2007). Xie-Beni validity index has also been chosen as the cluster validity measure because it has been shown to be able to detect the correct number of clusters in several experiments (Pal et. al. 1995).

K Means, Rough K-Means, Fuzzy C-Means and Penalized Fuzzy C-Means clustering algorithm are applied and analysed for Brain Tumour genes.

Table.1 shows the experimental results that are obtained by applying the above mentioned validity measures and the effectiveness of the proposed algorithm is well understood. We tested our method for the Brain Tumour gene expression dataset to cluster the highly suppressed and highly expressed genes and are depicted for various dataset sizes. It is observed that for each set of genes taken, the value of validity measures for the proposed algorithm is lower than the value of validity measures for other algorithms and it is graphically illustrated from Figure 1 to Figure 12. Among these clustering algorithms Penalized Fuzzy C-Means produces better results in identification of differences between data sets. This helps to correlate the samples according to the level of gene expression.

In terms of MAE, Penalized Fuzzy C-Means shows superior performance and K-Means and rough K-Means exhibits better performance than Fuzzy C-Means.

Also, With respect to RMSE and Xie-Beni Index Penalized Fuzzy C-Means produces greater performance than the other algorithms.

The comparative results based on Root Mean Square Error, Mean Absolute Error and Xie-Beni validity measure for all the Gene expression data clustering algorithms for various Brain Tumor data sets taken with different number of clusters are enumerated in Table 1.

Figure 1 to Figure 3 shows the comparative analysis of various approaches for 7129 genes taking K as 7, 5 and 3 respectively. It can be observed from the figures, Penalized Fuzzy C-Means outperforms other approaches Fuzzy C-Means, Rough K-Means and K-Means.

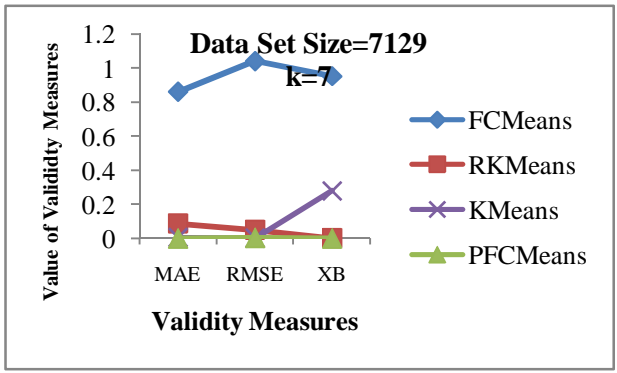

**Figure 1:** Validity Measure for Data set size = 7129, k=7

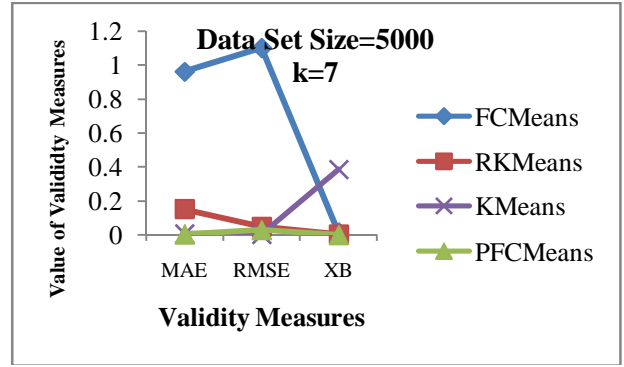

**Figure4:** Validity Measure for Data set size =5000, k=7

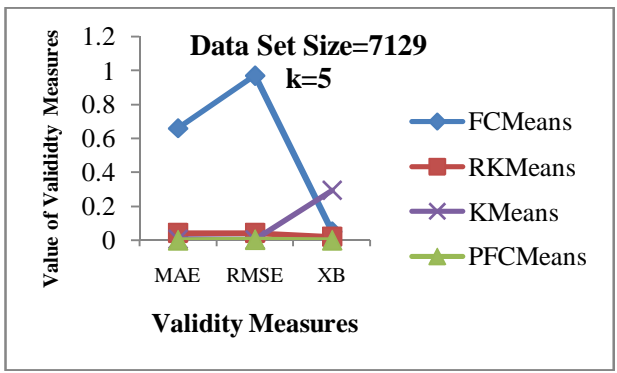

**Figure 2:** Validity Measure for Data set size =7129, k=5

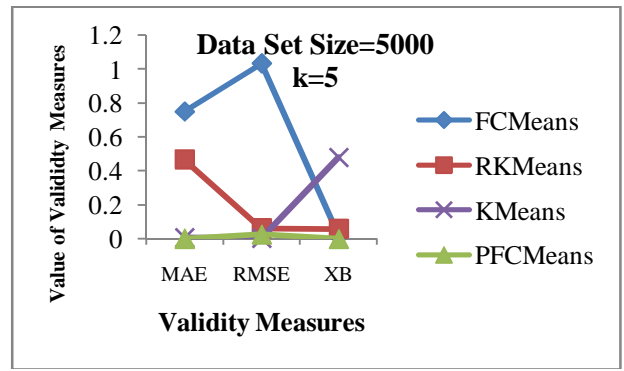

**Figure5:** Validity Measure for Data set size = 5000, k=5

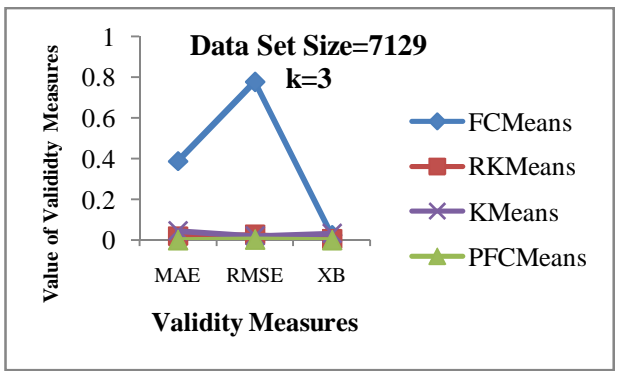

**Figure 3:** Validity Measure for Data set size = 7129, k=3

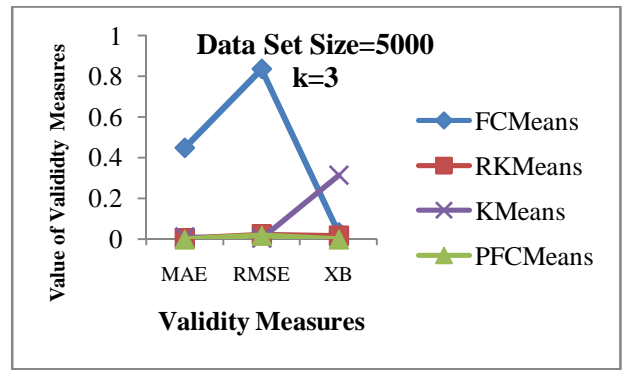

**Figure6:** Validity Measure for Data set size =5000, k=3

Figure 4 to Figure 6 shows the comparative analysis of various approaches for 5000 genes taking K as 7, 5 and 3 respectively. The experimental result shows that the Penalized Fuzzy C-Means outperforms other approaches Fuzzy C-Means, Rough K-Means and K-Means.

Figure 7 to Figure 9 shows the comparative analysis of various approaches for 3000 genes taking K as 7, 5 and 3 respectively. The experimental result shows that the Penalized Fuzzy C-Means gives better results than the other algorithms.

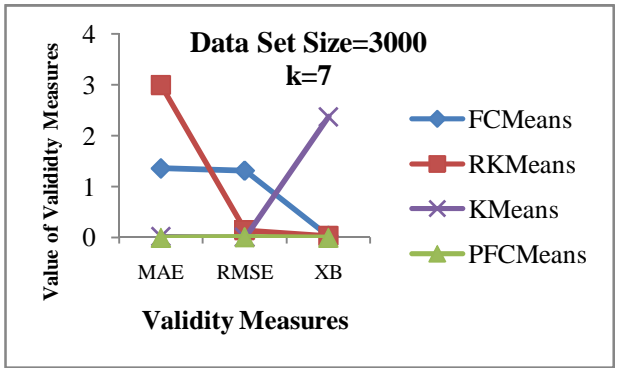

Figure 7: Validity Measure for Data set size = 3000, k=7

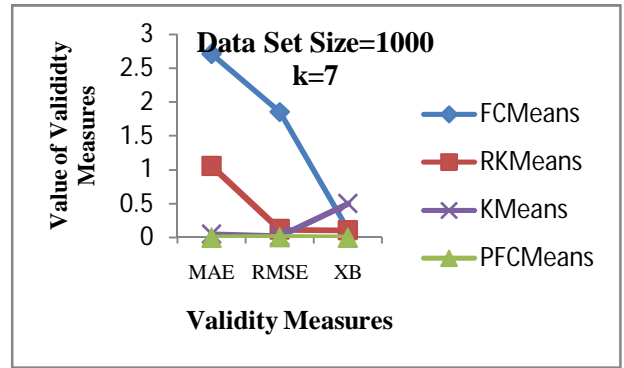

Figure10: Validity Measure for Data set size =1000, k=7

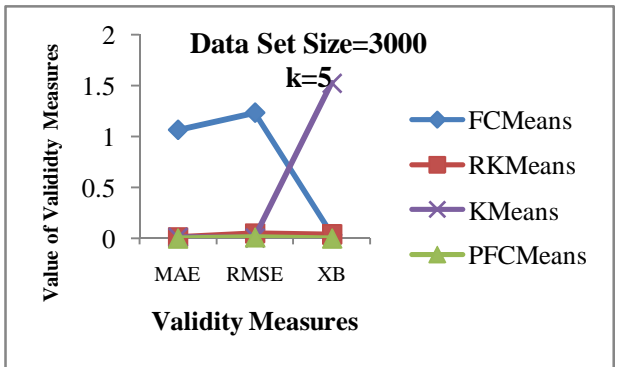

Figure8: Validity Measure for Data set size =3000, k=5

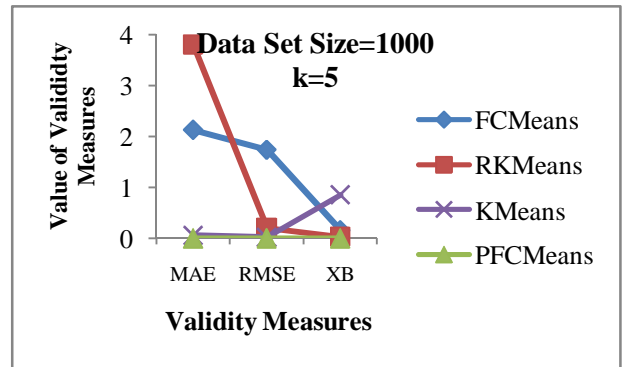

Figure 11: Validity Measure for Dataset Size =1000, k=5

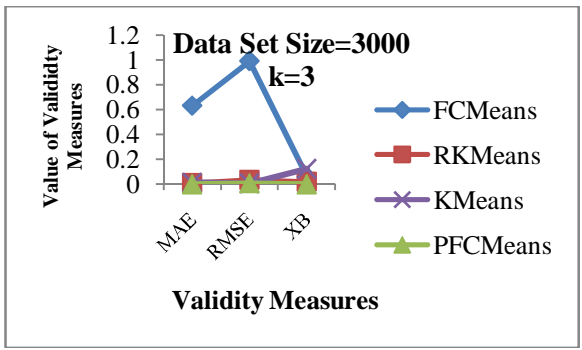

Figure 9: Validity Measure for Data set size = 3000, k=3

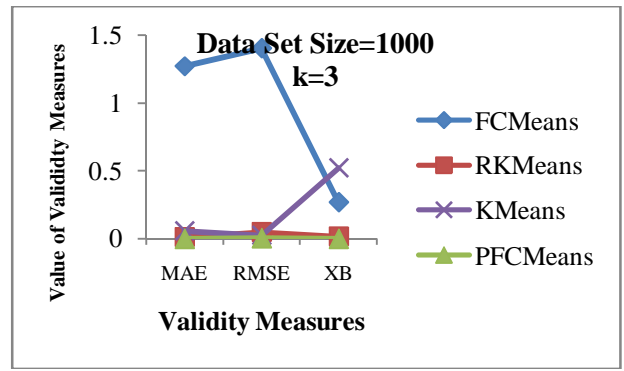

Figure12: Validity Measure for Data set size =1000, k=3

Figure 10 to Figure 12 shows the comparative analysis of various approaches for 1000 genes taking K as 7, 5 and 3 respectively. The experimental result shows that the Penalized Fuzzy C-Means performs better than the other algorithms

**Table 1: Performance Analysis of K-Means, Rough K-Means, Fuzzy C-Means and Penalized Fuzzy C-Means**

| No. of Clusters | No. of Genes | Clustering Algorithms | Root Mean Square Error | Mean Absolute Error | Xie-Beni Index |
|---|---|---|---|---|---|
| 7 | 7129 | K-Means | 0.0019 | 0.0044 | 0.2804 |
| | | Rough K-Means | 0.0511 | 0.0874 | 0.0020 |
| | | **Fuzzy C-Means** | **1.0438** | **0.8627** | **0.9548** |
| | | **Penalized Fuzzy C-Means** | **0.0043** | **0.0009** | **0.0001** |
| 5 | 5000 | K-Means | 0.0034 | 0.0074 | 0.4798 |
| | | Rough K-Means | 0.0624 | 0.4654 | 0.0581 |
| | | **Fuzzy C-Means** | **1.0336** | **0.7504** | **0.0184** |
| | | **Penalized Fuzzy C-Means** | **0.0255** | **0.0015** | **0.0002** |
| 3 | 3000 | K-Means | 0.0065 | 0.0150 | 0.1240 |
| | | Rough K-Means | 0.0336 | 0.0082 | 0.0203 |
| | | **Fuzzy C-Means** | **0.9925** | **0.6331** | **0.0624** |
| | | **Penalized Fuzzy C-Means** | **0.0079** | **0.0002** | **0.0001** |
| 7 | 1000 | K-Means | 0.0225 | 0.0470 | 0.4994 |
| | | Rough K-Means | 0.1204 | 1.0548 | 0.1079 |
| | | **Fuzzy C-Means** | **1.8542** | **2.7108** | **0.1107** |
| | | **Penalized Fuzzy C-Means** | **0.0056** | **0.0015** | **0.0001** |

**Table .2 Parameter setting and other issues**

| Clustering Algorithm | Cluster Membership | Input | Proximity Measure | Other Issues |
|---|---|---|---|---|
| K-Means | Binary | Starting Prototype, Stopping Threshold, K | Pair wise Distance | Very Sensitive to Input parameters and order of Input |
| Rough K-Means | Rough Membership | Starting Prototype, Stopping Threshold, K | Pair wise Distance | Imprecision in Gene Expression data can be captured |
| Fuzzy C-Means | Fuzzy Membership | Degree of fuzziness, Starting Prototypes, Stopping Threshold, K | Pair wise Distance | Careful interpretations of membership values. Sensitive to Input parameters and order of Input |
| **Penalized Fuzzy C-Means** | Improved Fuzzy Membership | Fuzzification Parameter, K | Pair wise Distance | Quality of membership is increased by introducing Penalty term |

*Pattern Analysis*

The parameter setting and other issues of the clustering approaches which are discussed are given in Table.2.

Transcriptional Initiation is the most important mode for control of gene expression level. Suppressed gene expression level may be stimulated by gene therapy (i.e.) promoter insertion or up regulation of suppressed gene and radiation therapy (i.e.) more amount of radiation may cause sudden mutation in gene, due to this sudden change the gene expression may be in high level. This helps to correlate the samples according to the level of gene expression. When precise functions for over or under expressed genes are determined, new avenues for intervention strategies may emerge. These studies are in their infancy; however, the improved technology employed here shows reasonable promise as our understanding of these deadly tumours increases. Therefore, future treatment decisions based on the expression profile of a primary tumour is a rational approach towards preventing the outgrowth of metastases.

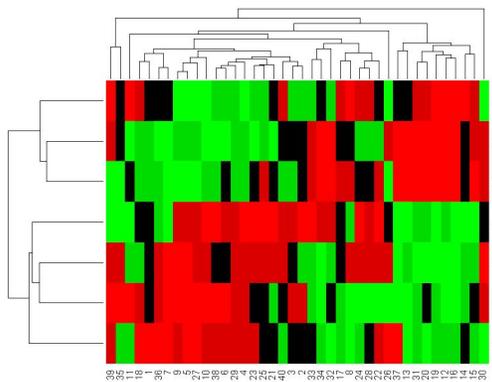

**Figure 13: Clusters of Brain Tumor Genes**

Figure.13 represents the Clusters of tumor genes by Penalized Fuzzy C-Means using all genes exhibiting variation across the data set.

For each gene, red indicates a high level of expression (highly expressed Genes) relative to the mean; green indicates a low level of expression (highly suppressed Genes) relative to the mean.

## 6. CONCLUSION

Cluster analysis applied to microarray measurements aims to highlight meaningful patterns for gene expression. The goal of gene clustering is to identify the important gene markers. Gene expression data are the representation of nonlinear interactions among genes and environmental factors. Brain Tumor is so deadly because, it is not usually diagnosed until it has reached an advance stage. Early detection can help prolong or save lives, but clinicians currently have no specific and sensitive method. Computing analysis of these data is expected to gain knowledge of gene functions and disease mechanisms. We used a set of gene expression data that contains a series of gene expression measurements of the transcript (mRNA) levels of Brain Tumor gene. Highly expressed genes and suppressed genes are identified and clustered using various clustering techniques. The following methods such as Post-Translational Modification, Small RNAs and Control of Transcript Levels, Translational Initiation, Transcript Stability, RNA Transport, Transcript Processing and Modification, Epigenetic Control and Transcriptional Initiation can be used to reduce the tumor level.

The empirical results also reveal the importance of using Penalized Fuzzy C-Means (PFCM) clustering methods for clustering more meaningful highly suppressed and highly expressed genes from the gene expression data set. Meanwhile, evaluation metrics Root Mean Square Error,

Mean Absolute Error and Xie-Beni Index is adopted to assess the quality of clusters, and the experimental results have shown that the proposed approach is capable of effectively discovering gene expression patterns and revealing the underlying relationships among genes as well. This clustering approach can be applied for any gene expression dataset and tumor growth can be predicted easily.